\documentclass[letterpaper, 10 pt, conference]{ieeeconf}  % Comment this line out if you need a4paper
\IEEEoverridecommandlockouts                              % This command is only needed if
\usepackage{cite}
\usepackage{amsmath,amssymb,amsfonts}

\usepackage{amsthm}
\usepackage{algorithmic}
\usepackage{graphicx}
\usepackage{epstopdf}
\usepackage{color}
\usepackage{textcomp}
\usepackage{dsfont}
\usepackage{verbatim}
\usepackage[linesnumbered,ruled]{algorithm2e}
\usepackage{hyperref}

\newtheorem{theorem}{Theorem}

\newcommand{\argmin}{\operatornamewithlimits{argmin}}
\newcommand{\R}{\mathbb{R}}

\newcommand{\Xext}{\mathbb{X}_{\mathrm{ext}}}

\definecolor{mygray}{gray}{0.6}
\title{ 
An Inverse Dynamics Approach to Control Lyapunov Functions
}
\author{Jenna Reher, Claudia Kann, and Aaron D. Ames$^{1}$% <-this % stops a space
\thanks{*This research is supported by the NSF Graduate Research Fellowship No. DGE‐1745301, under NSF Grant Numbers 1544332, 1724457, 1724464 and Disney Research LA.}% <-this % stops a space
\thanks{${^1}$J. Reher, C. Kann, and A. D. Ames are with the Department of Mechanical and Civil Engineering, Caltech, Pasadena, CA 91125 USA.}
}

\begin{document}

\maketitle
\thispagestyle{empty}
\pagestyle{empty}

%%%%%%%%%%%%%%%%%%%%%%%%%%%%%%%%%%%
%% Main Body
%%%%%%%%%%%%%%%%%%%%%%%%%%%%%%%%%%% 
%%%%%%%%%%%%%%%%%%%%%%%%%%%%%%%%%%%%%%%%%%%%%%%%%%%%%%%%
%% Abstract
%%%%%%%%%%%%%%%%%%%%%%%%%%%%%%%%%%%%%%%%%%%%%%%%%%%%%%%%
\begin{abstract}
    With the goal of moving towards implementation of increasingly dynamic behaviors on underactuated systems, this paper presents an optimization-based approach for solving full-body dynamics based controllers on underactuated bipedal robots. The primary focus of this paper is on the development of an alternative approach to the implementation of controllers utilizing control Lyapunov function based quadratic programs. This approach utilizes many of the desirable aspects from successful inverse dynamics based controllers in the literature, while also incorporating a variant of control Lyapunov functions that renders better convergence in the context of tracking outputs. The principal benefits of this formulation include a greater ability to add costs which regulate the resulting behavior of the robot. In addition, the model error-prone inertia matrix is used only once, in a non-inverted form. The result is a successful demonstration of the controller for walking in simulation, and applied on hardware in real-time for dynamic crouching. 
\end{abstract}

%%%%%%%%%%%%%%%%%%%%%%%%%%%%%%%%%%%%%%%%%%%%%%%%%%%%%%%%
%% Introduction
%%%%%%%%%%%%%%%%%%%%%%%%%%%%%%%%%%%%%%%%%%%%%%%%%%%%%%%%
\section{Introduction}
\label{sec:intro}
Model based control methods can help enable dynamic and compliant motion of robots while achieving remarkable control accuracy. However, implementing such techniques on floating base robots is non-trivial due to model inaccuracy, underactuation, dynamically changing contact constraints, and possibly conflicting objectives for the robot \cite{abe2007multiobjective,ames2013towards}. Unlike their classical counterparts, optimization based approaches to handling these control problems allow for the inclusion of physical constraints that the system is subject to \cite{posa2014direct, betts2002practical}. Partially as a consequence of this feature, quadratic programming (QP) based controllers have been increasingly used to stabilize real-world systems on complex robotic platforms without the need to algebraically produce a control law or enforce convergence guarantees \cite{koolen2016design, herzog2016momentum, feng2015optimization}.

These examples, however, typically do not consider periodic notions of stability for highly underactuated systems; systems which often require additional convergence guarantees in order to realize stability. It was shown in \cite{ames2014rapidly} that through the use of a \textit{rapidly exponentially stable control Lyapunov function} (RES-CLF), coupled with \textit{hybrid zero dynamics} (HZD) \cite{westervelt2018feedback, Grizzle2014Models}, a wide class of controllers can be designed to create rapidly exponentially convergent hybrid periodic orbits. It was also shown that this class of controllers can be posed as a QP, in which the convergence is enforced via an inequality constraint; forming a control Lyapunov function based quadratic program (CLF-QP) \cite{ames2014rapidly} \cite{ames2013towards}. Often, robotic systems cannot produce sufficient convergence to dynamic motions without violating physical constraints. One approach to address this conflict is to relax convergence guarantees, which allows (local) drift in the control objectives to accommodate feasibility. This class of controllers has since been used to achieve dynamic locomotion on robotic systems both in simulation \cite{nguyen2015optimal, kolathaya2016time, hereid2014embedding, xiong2018coupling} and on hardware \cite{galloway2015torque}. 

\begin{figure}[!t]
\centering
    \includegraphics[width = 0.97 \columnwidth]{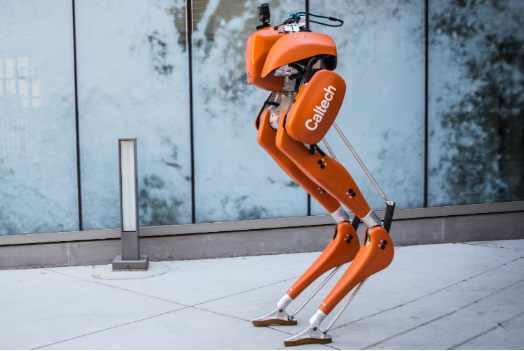}
    \caption{The Cassie biped, built by Agility Robotics, and used as an experimental platform to demonstrate the controllers presented in this work.}
	\label{fig:cover}
	\vspace{-6.mm}
\end{figure}

While high level task-space controllers based on inverse dynamics approaches pose similar problems as CLF-QPs, they have traditionally not been formulated in the same way. In implementations of CLF-QPs the vector fields associated with robotic systems are typically utilized, which involves costly computations. Alternatively, in task based controllers, the dynamics are an equality constraint. Here, objectives are driven towards their targets through PD controllers in the cost \cite{feng2015optimization}. There have been several connections shown in related research \cite{kuindersma2014efficiently, kuindersma2016optimization}, where control Lyapunov functions were included in an inverse dynamics controller via an LQR in the cost. In this work we aim to repurpose several of the more mature concepts from inverse dynamics based approaches and demonstrate a more efficient CLF inspired formulation.

The main result of this paper is an optimization-based control framework that couples convergence constraints from CLF-QPs with concepts from inverse dynamics based controllers. The presented controllers documented and available as part of a C++ open-source repository \cite{papergithub}. 
We begin in Section \ref{sec:background} by the CLF framework which yields rapid convergence. 
%for dynamic motions on underactuated systems. 
This is followed by Section \ref{sec:oldcontrol} which explores existing optimization based techniques for control. Section \ref{sec:control} details a new class of optimization based controllers based on the CLF construction. Section \ref{sec:implementableMethods} discusses how to apply these methods practically. In Section \ref{sec:results} the model for the bipedal robot Cassie is shown and the controller is demonstrated in both simulation for walking, and in real-time on hardware for dynamic crouching.
\section{Preliminaries on Control Lyapunov Functions}
\label{sec:background}

In classical nonlinear control design, analysis is typically performed on a dynamical system of the form:
\begin{align}
    \dot{x} = f(x) + g(x) u, \label{eq:nl_eom}
\end{align}
where $x \in X \subseteq \R ^n$ is the set of controllable states and $u \in U \subseteq \R^m$ is the control input. The mappings $f:\mathbb{R}^n \rightarrow \mathbb{R}^n$, $g:\mathbb{R}^n \rightarrow \mathbb{R}^{n \times m}$ are assumed to be locally Lipschitz continuous functions of $x$. Let us consider a feedback control system which tracks a set of  desired trajectories of the form:
\begin{align}
    y(x) = y^a(x) - y^d(\tau(x)),
\end{align}
where $y^a : X \rightarrow \R^m$ and $y^d : \R \times \R^a \rightarrow \R^m$ are smooth functions encoding the desired behavior to be realized via control. We assume that $y(x)$ has (vector) relative degree $r$ \cite{sastry2013nonlinear}. 
It is often the case in robotic systems that $r = 1$ if $y(x)$ depends on position and velocity and $r = 2$ if it only depends on position, i.e., configuration variables.
Taking the derivatives of the outputs along $f(x)$ and $g(x)$ we obtain,
\begin{align}
    y^{(r)}(x) = L_f^{(r)}y(x) + \underbrace{L_gL_f^{(r - 1)}y(x)}_{A}u, \label{eq:FBLy}
\end{align}
where $A$ is called the \textit{decoupling matrix} which is invertible in the case of a (vector) relative degree. This implies that the system \eqref{eq:nl_eom} is feedback linearizable, and we can then prescribe the following control law:
\begin{align}
    u(x) = A^{-1}\Big(-L_f^{(r)}y(x) + v\Big) \Rightarrow y^{(r)} = v, \label{eq:FBL}
\end{align}
where $v$ is an auxiliary feedback control value. 

To motivate later constructions, we consider a mechanical system with configuration space $\mathcal{Q}$, (local) coordinates $q \in \mathcal{Q}$, and states $x = (q^T,\dot{q}^T)^T \in T\mathcal{Q} = X$. Suppose that for \eqref{eq:nl_eom} there is a set of outputs $y(x) = (y_1(q,\dot{q})^T,y_2(q)^T)^T$ of vector relative degree $1$ and $2$, respectively, on a region of interest; that is for $y_1(q,\dot{q}) \in \R^{m_1}$ and $y_2(q) \in \R^{m_2}$ with $m = m_1 + m_2$ we assume the vector relative degree is 1 for $y_1$ and the $2$ for $y_2$, i.e., $(1, \ldots, 1, 2, \ldots, 2)$ with 1's appearing $m_1$ times and 2's appearing $m_2$ times. We can then write an output tracking problem:
\begin{align}
    y_1(q, \dot{q}, t) &= y_1^a(q, \dot{q}) - y_1^d(\tau(t,q)) \label{eq:reldeg1} \\
    y_2(q, t) &= y_2^a(q) - y_2^d(\tau(t,q)), \label{eq:reldeg2}
\end{align}
where $y^a$ and $y^d$ are the actual and desired outputs, and $\tau(t,q)$ is some parameterization of time for the desired outputs to evolve on. Assuming that the preliminary feedback \eqref{eq:FBL} has been applied to \eqref{eq:nl_eom}, we will render a linear system for the output dynamics with coordinates $\eta := (y_1^T, y_2^T, \dot{y}_2^T)^T$,
\begin{align}
    \dot{\eta} &= \begin{bmatrix} \dot{y}_1 \\ \dot{y}_2 \\ \ddot{y}_2 \end{bmatrix} = \underset{F}{\underbrace{\begin{bmatrix}  0 & 0 \\ 0 & \mathbf{I} \\ 0 & 0 \end{bmatrix}}} \eta + \underset{G}{\underbrace{\begin{bmatrix} \mathbf{I} & 0 \\ 0 & 0 \\ 0 & \mathbf{I} \end{bmatrix}}} v. \label{eq:output_dynamics}
\end{align}
A valid choice of $v$ which stabilizes this system is: 
\begin{align}
    v = \begin{bmatrix} \dot{y}_1 \\ \ddot{y}_2 \end{bmatrix} = \begin{bmatrix} - \frac{1}{\epsilon} K_{\bar{v}} y_1 \\ - \frac{1}{\epsilon^2} K_P y_2 - \frac{1}{\epsilon} K_D \dot{y}_2  \end{bmatrix}, \label{eq:IO_auxcontroller}
\end{align}
where $0 < \epsilon \leq 1$ is a tunable parameter, and $K_{\bar{v}}$, $K_P$, $K_D$ are control gains for the relative degree $1$ and relative degree $2$ output errors, respectively. While this controller yields convergence to the target outputs, it does not leverage the natural dynamics of the system, and disregards torque and feasibility constraints by which the system must abide. Thus, for practical systems, additional considerations for selecting our control input are often required.

The \textit{exponentially stabilizing control Lyapunov function} \cite{ames2012control} (ES-CLF) and \textit{rapidly exponentially stabilizing control Lyapunov function} (RES-CLF) frameworks \cite{ames2014rapidly} were introduced as methods for achieving stability in the output dynamics. 
In the context of the control system \eqref{eq:output_dynamics}, we consider the continuous time algebraic Riccati equations (CARE):
\begin{align}
    F^T P + P F - P G G^T P + Q = 0,
\end{align}
for $Q=Q^T > 0$ and with solution $P=P^T > 0$. The method presented in \cite{ames2014rapidly} can then be employed to construct a (R)ES-CLF, 
\begin{align}
    V(\eta) = \eta^T \underset{P_\epsilon}{\underbrace{\mathbf{I}_\epsilon P \mathbf{I}_\epsilon} \eta}, 
    \hspace{12pt} \mathrm{ with } \hspace{3 pt} \mathbf{I}_\epsilon := \mathrm{diag}\left( \mathbf{I},  \frac{1}{\epsilon} \mathbf{I}, \mathbf{I} \right), \label{eq:lyap}
\end{align}
where the selection of $0 < \epsilon < 1$ creates a RES-CLF, and $\epsilon = 1$ instead renders an ES-CLF. We can find the derivative of \eqref{eq:lyap} to be:
\begin{align}
    \dot{V}(\eta) &= L_F V(\eta) + L_G V(\eta) v, \label{eq:Vdot}
\end{align}
where the Lie derivatives of $V_\epsilon$ along the linear output system's dynamics \eqref{eq:output_dynamics} are
\begin{align}
    L_F V(\eta) &= \eta^T (F^T P_\epsilon + P_\epsilon F)\eta,\\
    L_G V(\eta) &= 2 \eta^T P_\epsilon G.
\end{align}
An exponential convergence constraint can then be prescribed as,
\begin{align}
    L_F V(\eta) + L_G V(\eta) v \leq - \underbrace{\frac{\lambda_{\mathrm{min}}(Q)}{\epsilon \lambda_{\mathrm{max}}(P_\epsilon)}}_{\gamma} V(x), \label{eq:CLFv}
\end{align}
where $\gamma$ is related to the convergence rate. This constraint is in terms of our auxiliary control input $v$ and not the actual feedback control $u$. In order to convert back into a form which can be represented in terms of the control input, we can use the previous relationship between $u$ and $v$
\begin{align}
    A(x) u + L_f ^{(r)}y(x) = v, \label{eq:acc_subst}
\end{align}
to obtain the CLF constraint stated in terms of $x$ since $\eta$ depends on $x$ (via $y_1$, $y_2$ and $\dot{y}_2$):
\begin{align}
    \underbrace{L_F V(x) + L_G V(x) L_f ^{(r)}y(x)}_{L_f V(x)} + \underbrace{L_G V(x) A(x)}_{L_g V(x)} u \leq - \gamma V(x).
    \label{eq:tradCLFconst}
\end{align}
In the context of (R)ES-CLF, we can then define the set
\begin{align}
    K(x) = \{u \in U : L_f V(x) + L_g V(x) u + \gamma V(x) \leq 0 \}, \label{eq:ES_clf_u_class}
\end{align}
consisting of the control values which result in (rapidly) exponential convergence, wherein $\dot{V}(\eta(x)) \leq -\gamma V(\eta(x))$.

\section{Optimization Based Controllers}
\label{sec:oldcontrol}
The dynamics of robotic systems can be formulated using the method of Lagrange, with positional constraints on the system incorporated via D'Alembert's principle \cite{Murray1994mathematical},
\begin{align}
    D(q)\ddot{q} + H(q, \dot{q}) &= Bu + J^T(q) \lambda \label{eq:eom} \\
    J(q) \ddot{q} + \dot{J}(q, \dot{q}) \dot{q} &= 0, \label{eq:Jeom}
\end{align}
where $D(q)$ is the inertia matrix, $H(q, \dot{q}) = C(q, \dot{q}) \dot{q} + G(q) + F$ is the vector sum for the Coriolis, centripital, gravitational, and additional non-conservative generalized forces, $B$ is the actuation matrix, and the Jacobian of the holonomic constraints is $J(q) = \partial h/ \partial q $ with its corresponding constraint wrenches $\lambda \in \mathbb{R}^{m_h}$. 
This can be converted to an ODE in the form of \eqref{eq:nl_eom} as:
\begin{align}
    f(x) &= \begin{bmatrix} \dot{q} \\ - D^{-1}(q) \left( J^T(q) \lambda - H(q,\dot{q}) \right) \end{bmatrix}, \nonumber \\
    g(x) &= \begin{bmatrix} 0 \\ D(q)^{-1} B \end{bmatrix}. \label{eq:manip_nl_eom}
\end{align}
We begin the derivation of our controller for this system by again considering the outputs \eqref{eq:reldeg1} and \eqref{eq:reldeg2}, and taking the necessary derivatives of the outputs:
\begin{align}
    \begin{bmatrix} \dot{y}_1 \\ \ddot{y}_2 \end{bmatrix} &= \underbrace{\begin{bmatrix} \frac{\partial y_1}{\partial q} \\ \frac{\partial }{\partial q} \left(\frac{\partial y_2}{\partial q}\dot{q} \right) \end{bmatrix}}_{\dot{J}_y} \dot{q} + \underbrace{\begin{bmatrix} \frac{\partial y_1}{\partial \dot{q}} \\ \frac{\partial y_2}{\partial q} \end{bmatrix}}_{J_y} \ddot{q}. \label{eq:ddqexplicit}
\end{align}
This can equivalently be done by taking the derivatives along the vector fields \eqref{eq:manip_nl_eom}, where the dependencies have been dropped for the sake of clarity, we can write:
\begin{align}
    \begin{bmatrix} \dot{y}_1 \\ \ddot{y}_2 \end{bmatrix} &= \underbrace{\begin{bmatrix} \frac{\partial y_1}{\partial q} & \frac{\partial y_1}{\partial \dot{q}} \\ \frac{\partial }{\partial q} \left(\frac{\partial y_2}{\partial q}\dot{q} \right) & \frac{\partial y_2}{\partial q} \end{bmatrix} f(x)}_{L_fy(x)} + \underbrace{\begin{bmatrix} \frac{\partial y_1}{\partial \dot{q}} \\ \frac{\partial y_2}{\partial q} \end{bmatrix} g(x)}_{A} u,
\end{align}
now in the form of \eqref{eq:FBL}. This can then be combined with the convergence constraint given for a CLF \eqref{eq:tradCLFconst} and posed as an optimization problem to find a satisfactory input $u$.
 
\vspace{2mm} \noindent 
\textbf{Control Lypaunov Function Quadratic Programs:}
In its traditional implementation \cite{ames2014rapidly}, the inequality constraint \eqref{eq:tradCLFconst} can be posed in a QP optimization based controller to find a torque in the set \eqref{eq:ES_clf_u_class}, where $\|v\|^2$ is minimized, as:

\noindent\rule{\columnwidth}{1pt}
\textbf{CLF-QP:}
{\footnotesize
\begin{align*}
    u^{\ast}(x) = \argmin_{u\in U \subset \mathbb{R}^m} \quad & ||A(x)u + L_f^{(r)}y(x)||^2\\
    \mathrm{s.t.} \quad & L_f V(x) + L_g V(x)u \leq \gamma V(x)
\end{align*}}%
\noindent\rule{\columnwidth}{1pt}
Where in the case of a RES-CLF, $\gamma$ depends on $\epsilon$. For the holonomic constraints to be satisfied in the dynamics \eqref{eq:manip_nl_eom}, and thus in the QP constraint \eqref{eq:tradCLFconst}, we must either augment $u$ with $\lambda$ as an additional decision variable \cite{hereid2014embedding, ames2013towards}, or solve for the generalized force explicitly, assuming the holonomic constraint is satisfied:
 \begin{align}
    \lambda &= (J_c D^{-1} J^T_c)^{-1} \left( J_c D^{-1} (H - B u) - \dot{J}_c \dot{q} \right), \label{eq:solveF}
\end{align}
and substitute back into the expression \eqref{eq:manip_nl_eom}. 
However, even if $\lambda$ is included as an additional optimization variable, \eqref{eq:solveF} must be evaluated in order to apply feasibility constraints such as the friction cone to the problem. Additionally, Featherstone showed in previous work that the condition number of the joint space inertia matrix increases quartically with the length of a kinematic chain \cite{featherstone2004empirical}. This points to an obvious source of numerical stiffness, and can lead to controller degradation on hardware \cite{nakanishi2008operational}. For complex multi-link robots, such as bipedal robots, these condition numbers are often exceptionally large (for full humanoids sometimes on the order of $10^8$). In addition, performing the required inversions for evaluating the vector fields \eqref{eq:manip_nl_eom} are very computationally expensive, and can often violate strict timing requirements when implementing these controllers on hardware.

%%%%%%%%%%%%%%%%%%%%%%%%%%%%%%%%%%%%%%%%%%%%%%%%%%%%%%%%%%%%%%%%%%%%%%%%%%%%%%%%%%%%%%%%%%%%%%%%%%%%%%%%
%%%%%%%%%%%%%%%%%%%%%%%%%%%%%%%%%%%%%%%%%%%%%%%%%%%%%%%%%%%%%%%%%%%%%%%%%%%%%%%%%%%%%%%%%%%%%%%%%%%%%%%%
\vspace{2mm} \noindent
\textbf{Inverse Dynamics Approaches to Locomotion:}
\textit{Inverse dynamics} is a widely used method to approaching controller design for achieving a variety of motions and force interactions, typically in the form of task-space objectives. Given a target behavior, the dynamics of the robotic system are inverted to obtain the desired torques. In most formulations, the system dynamics are mapped onto a support-consistent manifold using methods such as the dynamically consistent support null-space \cite{sentis2007synthesis}, linear projection \cite{aghili2005unified}, and orthogonal projection \cite{mistry2010inverse}. When prescribing behaviors in terms of purely task space objectives, this is commonly referred to as task- or operational-space control (OSC) \cite{khatib1987unified}. In many recent works, variations of these approaches have been shown to allow for high-level tasks to be encoded with intuitive constraints and costs in optimization based controllers, some examples being \cite{hutter2014quadrupedal, apgar2018fast, kuindersma2016optimization, feng2015optimization, koolen2016design, herzog2016momentum, kuindersma2014efficiently}.

Here we present a minimal implementation of an inverse dynamics controller. First, let us consider a set of variables $\mathcal{X} = [ \ddot{q}^T, u^T, \lambda^T]^T \in \Xext := \R^n \times U \times \R^{m_h}$, which are linear with respect to \eqref{eq:eom},
\begin{align}
    \begin{bmatrix} D(q) & -B & - J^T(q) \end{bmatrix} \mathcal{X} + H(q,\dot{q}) = 0. \label{eq:IDform}
\end{align}
We can pose the holonomic constraints \eqref{eq:Jeom} as:
\begin{align}
    \begin{bmatrix} J(q) & 0 & 0 \end{bmatrix} \mathcal{X} + \dot{J}(q)\dot{q} = 0.
\end{align}
Also consider a positional objective in the task space of the robot, which can be characterized using \eqref{eq:ddqexplicit} as
\begin{align}
    J_y(q,\dot{q}) \ddot{q} + \dot{J}_y(q,\dot{q}) \dot{q} - \ddot{y}_2^* = 0,
\end{align}
where $\ddot{y}_2^* = K_P y_2 + K_D \dot{y}_2$ is a PD control law which can be tuned to achieve convergence. In it's most basic case, not considering physical limitations on torque and frictional contact, we can pose this tracking problem as:

\noindent\rule{\columnwidth}{1pt}
\textbf{ID-QP:}
{\footnotesize
\begin{align*}
    \mathcal{X}^{\ast}(x) = \argmin_{\mathcal{X}\in \Xext} \quad & ||\dot{J}_y(q,\dot{q}) \ddot{q} + J_y(q,\dot{q}) \dot{q} - \ddot{y}_2^* ||^2  + \sigma W(\mathcal{X})\\
   \textrm{ s.t.} \quad &D(q)\ddot{q} + H(q,\dot{q}) = Bu + J^T(q)\lambda\\
    & J(q)
    \ddot{q} + \dot{J}(q)\dot{q} = 0
    \vspace{-10mm}
\end{align*}}%
\noindent\rule{\columnwidth}{1pt}
where $W(\mathcal{X})$ is included as a regularization term with a small weight $\sigma$ such that the problem is well posed.

\section{Controller Formulation}
\label{sec:control}

In this section, a new controller is presented that combines aspects of ID and CLF based control. The resulting optimization only requires a single use of the mass matrix, in its uninverted form, and incentivizes fast convergence rates.

\vspace{2mm} \noindent
\textbf{A Combined Approach.}
Taking inspiration from inverse dynamics approaches, we return to \eqref{eq:acc_subst} where the auxiliary control input $v$ is set to equal the second time derivative of the output. Rather than directly choosing an input $u$, a $\ddot{q}$ is  solved for that generates an equivalent response in the outputs. 
%%%%%%%%%%%%%%%%%%%%%%%%%%%%%%%%%%%%%%%%%
%%%%% NEW POSSIBLY TO DELETE %%%%%%%%%%%%
Using \eqref{eq:ddqexplicit}, $\ddot{q}$ can be chosen to satisfy
\begin{align}
    \begin{bmatrix} \dot{y}_1 \\ \ddot{y}_2 \end{bmatrix} &= \dot{J}_y\dot{q} + J_y\ddot{q} = v.
\end{align}
By constraining:
\begin{align}
    \ddot{q} = J^{\dagger}_y(-\dot{J}_y\dot{q} + v), \label{eq:ddqFBL}
\end{align}
where $J^{\dagger}_y$ is a right pseudo inverse of the full rank matrix $J_y$, with $J_y J^{\dagger}_y = I$, and the outputs evolve as:
\begin{align}
    \begin{bmatrix} \dot{y}_1 \\ \ddot{y}_2 \end{bmatrix} &= \dot{J}_y\dot{q} + J_yJ^{\dagger}_y(-\dot{J}_y\dot{q} + v) = v.
\end{align}
More formally, we have shown the following result:

\begin{theorem}
\textit{For a robotic system with dynamics \eqref{eq:eom} and outputs of the form \eqref{eq:reldeg1} and \eqref{eq:reldeg2}, any controller in the set:
\begin{align}
    K(q,\dot{q}) = \{u\in U :    \ddot{q} =J^{\dagger}_y(-\dot{J}_y\dot{q} + v)  \},
\end{align}
elicits the same response in the output dynamics as the IO feedback linearizing controller,
\begin{align}
    u = A^{-1}(-L_fy(x)+v).
\end{align}
}
\end{theorem}

\vspace{0.3cm}

%%%%%%%%%%%%%%%%%%%%%%%%%%%%%%%%%%%%%%%%%
\vspace{-2mm}
As discussed in Section \ref{sec:background}, feedback linearizing controllers fail to take advantage of the natural dynamics of a system. Therefore, we introduce the \textit{Inverse Dynamics Control Lyapunov Function Quadratic Program} (\textbf{ID-CLF-QP}) a parallel to (\textbf{CLF-QP}), which similarly enables the system to evolve in a more natural way, while still enforcing convergence guarantees.

\noindent\rule{\columnwidth}{1pt}
\textbf{ID-CLF-QP}
{\footnotesize
\begin{align*}
    \mathcal{X}^{\ast} = \argmin_{\mathcal{X}\in \Xext} \quad & \bigg|\bigg|\frac{\partial\dot{y}}{\partial q}\dot{q} +  \frac{\partial y(q)}{\partial q}\ddot{q}\bigg|\bigg|^2+\sigma W(\mathcal{X})\\
    \textrm{s.t.} \quad &  L_F V(x) + L_G V(x) \Bigg(\frac{\partial\dot{y}}{\partial q}\dot{q} +  \frac{\partial y(q)}{\partial q}\ddot{q} \Bigg) \leq - \gamma V(x)\\
    &D(q)\ddot{q} + H(q,\dot{q}) = Bu + J^T(q)\lambda\\
    & J(q)
    \ddot{q} + \dot{J}(q)\dot{q} = 0
\end{align*} }%
\noindent\rule{\columnwidth}{1pt}%

This formulation imposes an equivalent convergence condition as (\textbf{CLF-QP}). However, using $\ddot{q}$ as an optimization variable leads to a formulation that is less numerically stiff and less sensitive to estimation errors in the mass matrix. 

\vspace{2mm} \noindent
\textbf{Incentivized Convergence.} A second weakness of the standard (\textbf{CLF-QP}) is that it does not incentivize faster convergence rates than the chosen $\gamma$ if control bandwidth is available. This lead to chattering as the system intermittently triggers the inequality. When the outputs are written as in \eqref{eq:ddqexplicit}, the derivative of the Lyapunov function is only in terms of the decision variable $\ddot{q}$ and scalar functions of the states. Therefore, we add the $\ddot{q}$-dependent portion to the cost.

\noindent\rule{\columnwidth}{1pt}
\textbf{ID-CLF-QP$^{+}$}
{\footnotesize
\begin{align*}
    \mathcal{X}^{\ast} = \argmin_{\mathcal{X}\in \Xext} \quad & \bigg|\bigg|\frac{\partial\dot{y}}{\partial q}\dot{q} +  \frac{\partial y(q)}{\partial q}\ddot{q}\bigg|\bigg|^2+\sigma W(\mathcal{X}) + \dot{V}(x,\mathcal{X}) \\ %L_G V(x)\frac{\partial y(q)}{\partial q}\ddot{q}  \\
    \textrm{s.t.} \quad & L_F V(x) + L_G V(x) \Bigg(\frac{\partial\dot{y}}{\partial q}\dot{q} +  \frac{\partial y(q)}{\partial q}\ddot{q} \Bigg) \leq - \gamma V(x)\\
    &D(q)\ddot{q} + H(q,\dot{q}) = Bu + J^T(q)\lambda\\
    & J(q)
    \ddot{q} + \dot{J}(q)\dot{q} = 0
\end{align*}}%
\noindent\rule{\columnwidth}{1pt}

\vspace{2 mm}

\begin{theorem}
\label{thm:main}
\textit{Through the addition of a Lyapunov term in the cost, (\textbf{ID-CLF-QP$^{+}$}) will induce an equal or faster convergence rate than (\textbf{ID-CLF-QP}).
Concretely, given solutions to these optimization problems, denoted by $\mathcal{X}^{+}$ and $\tilde{\mathcal{X}}$, respectively, for:
$$
\begin{array}{rl}
\dot{V}(x,\mathcal{X}^{+}) & \leq   - \gamma^+ V(x)  \\
\dot{V}(x,\tilde{\mathcal{X}}) & \leq  - \tilde{\gamma} V(x)
\end{array}
\qquad \Rightarrow \qquad 
\tilde{\gamma} \leq \gamma^+.
$$
}
\end{theorem}

\vspace{2 mm}
\noindent\textit{Proof:} We begin by noting that (\textbf{ID-CLF-QP}) and (\textbf{ID-CLF-QP$^{+}$}) have the same solution space. The cost of (\textbf{ID-CLF-QP}) as will be denoted:
\begin{align}
    C(\mathcal{X}) =  \bigg|\bigg|\frac{\partial\dot{y}}{\partial q}\dot{q} +  \frac{\partial y(q)}{\partial q}\ddot{q}\bigg|\bigg|^2+\sigma W(\mathcal{X})
\end{align}For a given feasible point $\mathcal{X}^{\ast}$ of (\textbf{ID-CLF-QP$^{+}$}),  the unique instantaneous convergence rate $\gamma^{\ast}$ is defined as solving:
\begin{align}
    L_F V(x) + L_G V(x) \Bigg(\frac{\partial\dot{y}}{\partial q}\dot{q} +  \frac{\partial y(q)}{\partial q}\ddot{q}^{\ast} \Bigg) = - \gamma^{\ast} V(x), \label{eq:rate}
\end{align}
and the cost is given by:
\begin{align}
    \mathcal{J}(\mathcal{X}^{\ast}) = C(\mathcal{X}^{\ast}) + L_G V(x)\frac{\partial y(q)}{\partial q}\ddot{q}. 
\end{align}
Larger values of $\gamma$ imply a faster convergence rate. Denoting the solution to (\textbf{ID-CLF-QP}) as $\tilde{\mathcal{X}}$, then by definition:
\begin{align}
    C(\tilde{\mathcal{X}}) \leq C(\mathcal{X}^{\ast}), \quad \forall\mathcal{X}^{\ast} \in \mathcal{X}, \label{eq:bestX}
\end{align}
and additionally, for the solution to (\textbf{ID-CLF-QP$^{+}$}), $\mathcal{X}^+$ : 
\begin{align}
      \mathcal{J}(\mathcal{X}^{+}) \leq \mathcal{J}(\tilde{\mathcal{X}}), \label{eq:+cost}
\end{align}
as the two problems have the same feasible space. This can be expanded to:
\begin{align*}
   \underbrace{ L_G V(x)\frac{\partial y(q)}{\partial q}\ddot{q}^{+}+ C(\mathcal{X}^{+})}_{\mathcal{J}(\mathcal{X}^{+})}  \leq \underbrace{L_G V(x)\frac{\partial y(q)}{\partial q}\tilde{\ddot{q}}+ C(\tilde{\mathcal{X}})}_{\mathcal{J}(\tilde{\mathcal{X}})}.
\end{align*}
If we solve for the Lyapunov portion of each side using \eqref{eq:rate}, this reduces to:
\begin{align*}
    - \gamma^{+} V(x) + C(\mathcal{X}^{+}) \leq - \tilde{\gamma} V(x) + C(\tilde{\mathcal{X}}).
\end{align*}
This can be rearranged, and \eqref{eq:bestX} can be leveraged to find
\begin{align*}
    \tilde{\gamma} V(x) - \gamma^{+} V(x)  &\leq   C(\tilde{\mathcal{X}})-C(\mathcal{X}^{+}) \leq 0\\
    (\tilde{\gamma}  - \gamma^{+}) V(x)   &\leq 0\\
    \tilde{\gamma} &\leq \gamma^+.
\end{align*}
Which proves that the solution to (\textbf{ID-CLF-QP$^+$}) will have an equal or faster convergence rate as (\textbf{ID-CLF-QP}). \rule{0.7em}{0.7em}

\section{Implementable Methods}
\label{sec:implementableMethods}
In order to implement these methods on robotic systems, there must be a discussion on practicality and how to better encode and satisfy the physical limitations of the system at hand. In this section, some of the barriers to implementation are presented as well as methods for mitigating them.

%%%%%%%%%%%%%%%%%%%%%%%%%%%%%%%%%%%%%%%%%%%%%%%%%%%%%%%%
%% Holonomic Constraints
%%%%%%%%%%%%%%%%%%%%%%%%%%%%%%%%%%%%%%%%%%%%%%%%%%%%%%%%

\vspace{2mm} \noindent
\textbf{Holonomic Constraints.}
For robotic systems, two types of holonomic constraints are commonly considered, external contact constraints depending on the current configuration of the robot and it's interactions with the world, and internal kinematic constraints resulting from the robot geometry.

%%%%%%%%%%%%%%%%%%%%%%%%%%%%%%%%%%%%%%%%%%%%%%%%%%%%%%%%
%% Foot constraints
%%%%%%%%%%%%%%%%%%%%%%%%%%%%%%%%%%%%%%%%%%%%%%%%%%%%%%%%
\subsubsection{Contact Constraints}
When the robot is in contact with the world, its motion can be restricted. This results in force terms in the equations of motion (\eqref{eq:eom} and \eqref{eq:Jeom}). These contacts are often required to follow friction models. Ideally, a classical \textit{Amontons-Coulomb model} of (dry) friction is used to avoid slippage and is represented as a friction cone constraint. For a friction coefficient $\mu$ and a surface normal, the space of valid reaction forces is,

\vspace{-2mm}
{\small
\begin{align}
    \mathcal{C} = \left\{ \left. ( \lambda_x, \lambda_y, \lambda_z ) \in \R^3 \right| \lambda_z \geq 0; \sqrt{\lambda_x^2 + \lambda_y^2} \leq \mu \lambda_z \right\}. \label{eq:cone_friction}
\end{align}}
However, this constraint is nonlinear, and cannot be implemented as a linear constraint. % in our later formulations. 
An alternative solution is to use a \textit{pyramidal friction cone} approximation \cite{Grizzle2014Models},

\vspace{-2mm}
{\small
\begin{align}
    \mathcal{P} = \left\{ \left. ( \lambda_x, \lambda_y, \lambda_z )\in \R^3 \right| \lambda_z \geq 0; |\lambda_x|, |\lambda_y| \leq \frac{\mu}{\sqrt{2}} \lambda_z \right\}. \label{eq:pyramid_friction}
\end{align}}
This is a more conservative model than the friction cone, but is advantageous in that it is a linear inequality constraint. When a surface is in contact with the outside world, additional constraints are introduced to prevent it from rolling over the contact edge in the form:
\begin{align}
      -\frac{l}{2} \lambda_z <  &\lambda_{mx} < \frac{l}{2} \lambda_z  \label{eq:roll}\\
       -\frac{w}{2} \lambda_z <  &\lambda_{my} < \frac{w}{2} \lambda_z
\end{align}
where $l$ and $w$ are the lengths and widths of the surface.% \cite{vucobratovic1990biped}. 

%%%%%%%%%%%%%%%%%%%%%%%%%%%%%%%%%%%%%%%%%%%%%%%%%%%%%%%%
%% Loop constraints
%%%%%%%%%%%%%%%%%%%%%%%%%%%%%%%%%%%%%%%%%%%%%%%%%%%%%%%%
\subsubsection{Internal Constraints}
It is common practice to model robotic manipulators in tree structures. When the mechanism has parallel manipulators, this is managed by cutting the loop and enforcing a holonomic constraint \cite{reher2019dynamic}, or by solving for the closed-loop dynamics explicitly. These constraints add further degrees of complexity to the optimization problem.  

\vspace{2mm} \noindent
\textbf{Relaxed CLF-QP.}
Due to these constraints, as well as limits on feasible torques, it is not always possible for the system to converge according to the bound \cite{galloway2015torque}. The accepted way of dealing with this is to add a relaxation term, $\delta$, to the convergence constraint with an associated weight, $\rho$. In our formulation, this transforms the problem to:

\noindent\rule{\columnwidth}{1pt}
\textbf{ID-CLF-QP$^{+}$  $\delta$}
{\footnotesize
\begin{align*}
        \mathcal{X}^{\ast} = &\argmin_{\mathcal{X}\in \Xext, \delta \in \mathbb{R} } \quad  \bigg|\bigg|\frac{\partial\dot{y}}{\partial q}\dot{q} +  \frac{\partial y(q)}{\partial q}\ddot{q}\bigg|\bigg|^2+W(\mathcal{X}) + \dot{V}(x,\mathcal{X}) + \rho\delta^2 \\ %L_G V(x)\frac{\partial y(q)}{\partial q}\ddot{q} + \rho\delta^2\\
    & \textrm{s.t.} \quad   L_F V(x) + L_G V(x) \Bigg(\frac{\partial\dot{y}}{\partial q}\dot{q} +  \frac{\partial y(q)}{\partial q}\ddot{q} \Bigg) \leq - \gamma V(x)+ \delta\\
    & \hspace{.68cm} D(q)\ddot{q} + H(q,\dot{q}) = Bu + J^T(q)\lambda\\
    & \hspace{.68cm} J(q)
    \ddot{q} + \dot{J}(q)\dot{q} = 0
\end{align*}}
\noindent\rule{\columnwidth}{1pt}
In practice it can be seen that if we take away the hard constraint in (\textbf{ID-CLF-QP$^{+}$}), we are left with a relaxation 

\noindent\rule{\columnwidth}{1pt}
\textbf{ID-CLF-QP$^{+}$ relaxed}
{\footnotesize
\begin{align*}
        \mathcal{X}^{\ast}= &\argmin_{\mathcal{X}\in \Xext} \quad  \bigg|\bigg|\frac{\partial\dot{y}}{\partial q}\dot{q} +  \frac{\partial y(q)}{\partial q}\ddot{q}\bigg|\bigg|^2+W(\mathcal{X}) + \dot{V}(x,\mathcal{X}) \\ %+ L_G V(x)\frac{\partial y(q)}{\partial q}\ddot{q} \\
    & \textrm{s.t.} \quad   D(q)\ddot{q} + H(q,\dot{q}) = Bu + J^T(q)\lambda\\
    & \hspace{.68cm} J(q) \ddot{q} + \dot{J}(q)\dot{q} = 0 
\end{align*}}
\noindent\rule{\columnwidth}{1pt}
that still incentivizes fast convergence and penalizes slow convergence. Further, whenever it is feasible to do so, this problem will render $\dot{V}$ as negative as possible. In the simulation results we show how these methods compare and we implement the final approach on hardware.

\vspace{2mm} \noindent
\textbf{Hard and Soft Constraints.} When implementing on hardware, often holonomic constraints are not satisfied precisely. The analytical solutions presented thus far make the problem more prone to infeasibility. To solve this, we once again look to the inverse dynamics community where it has become practice to differentiate between \textit{hard} and \textit{soft} constraints. Hard constraints are formulated as traditionally seen in Section \ref{sec:control}, they cannot be violated. Soft constraints, however, refer to an addition to the cost function which penalizes violation of a preferred relationship. These are frequently added as the norm of a least squares problem:
\begin{align}
    w||\mathcal{A}\mathcal{X} - b||^2
\end{align}
where $\mathcal{X}$ is as in \eqref{eq:IDform} and $w$ is a weight. On hardware, holonomic constraints for footholds are the perfect candidate to be implemented as soft constraints
\begin{align}
    J(q)\ddot{q}+ \dot{J}\dot{q} = 0 \Rightarrow \underbrace{\begin{bmatrix} J(q) & 0 & 0\end{bmatrix}}_{\mathcal{A}}\mathcal{X} = \underbrace{-\dot{J}\dot{q}}_{b}
\end{align}
The formulation of holonomic constraints in this way voids the need to explicitly compute reaction forces (as in \eqref{eq:solveF}) and allows for small violations, which is necessary in practice. Additional soft constraints that are beneficial for robotic walking include specifying force distributions (weight per foot or in different places on the foot). In this case the $A$ matrix represents the fractional representation and the $b$ matrix is all zeros. Finally, direct tracking of decision variables is possible by making $\mathcal{A}$ the identity and $b$ the desired values. In each of these cases, a desirable cost is the exact satisfaction of the output dynamics. The benefit of using soft constraints is two-fold; as previously mentioned it allows for small violations of constraints and, it speeds up computation time as the problem becomes better posed.

\begin{figure}[!t]
\vspace{2mm}
	\centering
	\includegraphics[width= 1.00\columnwidth]{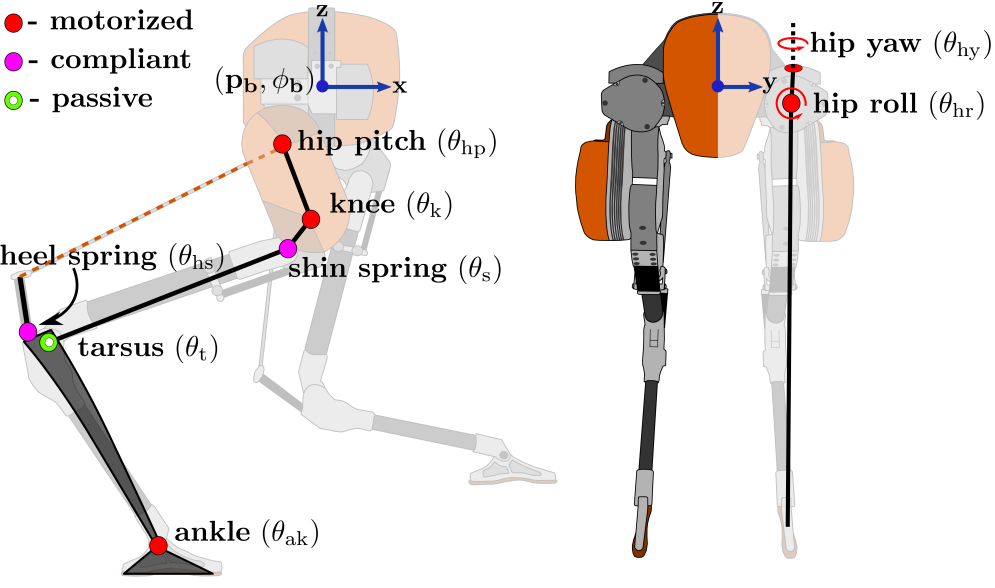}
	\caption{The configuration coordinates of the Cassie robot, on the left  is a side view of the robot, which highlights the compliant mechanism, and on the right is the front view of the robot model. 
    }
	\label{fig:config}
	\vspace{-6mm}
\end{figure}
\section{Application to the Cassie Biped}
\label{sec:results}

In this section the robot model will be introduced, followed by a presentation of simulation results for walking with each of the controllers mentioned, as well as real-time hardware results on crouching and standing behaviors.

\vspace{2mm} \noindent
\textbf{Robot Model.}
 The bipedal robot Cassie was designed and manufactured by Agility Robotics\footnote{http://www.agilityrobotics.com/}. 
The design of the robot encompasses the physical attributes of the spring loaded inverted pendulum (SLIP) model dynamics. 
 The primary characteristic being a pair of light-weight legs with a heavy torso so that the system is approximated by a point-mass with virtual springy legs. On Cassie, a \textit{compliant multi-link mechanism} is used to transfer power from higher to lower limbs without allocating the actuators' weight onto the lower limbs, and effectively acts as a pair of springy legs. 
Contacts with the ground are assumed to be rigid and only occurring at specified points on the feet of the robot. This allows for the equations of motion for the robot to be described as \eqref{eq:eom}.

\subsection{Walking in Simulation}
The simulation presents a side-by-side comparison of the traditional (\textbf{CLF-QP}) with the new controllers proposed. Two walking gaits are generated using the \textit{partial hybrid zero dynamic} framework as presented in \cite{westervelt2018feedback}. Both gaits use a single continuous domain, and progress is dictated by $\tau(t,q)$, a parameterization of time either by the gait duration (time-based) or by the relative degree $1$ output (state-based): 
\begin{align}
\tau(t) := \frac{t-t^1}{t^2-t^1} \quad \textrm{or}\quad  \tau(q) := \frac{\delta y_1(q) - \delta y_1(q^+)}{\bar{v}},
\end{align}
where  $t_0$ and $t_f$ are the start and end times of the current domain, respectively, $\delta y_1 (q^+)$ is the initial value of the velocity modulating output and $\bar{v}$ is a parameter for the desired velocity of the output. 

\vspace{2mm} \noindent
\textbf{Planar Walking Simulation.}
The first gait we consider is state-based and is designed on a planar, rigid model of Cassie. One relative degree 1 input, the linearized forward hip velocity, is used and five relative degree 2 outputs:

\begin{align}
    y_2^a(q) := \begin{bmatrix} 
    ||\psi^{sw}||_{\ell 2}\\
    ||\psi^{st}||_{\ell 2}\\
    \textrm{\footnotesize{atan2}}\big(\psi^{sw}_x,\psi^{sw}_z\big)\\
    \phi_y\\
    \phi^y(q)\end{bmatrix} \begin{pmatrix}
    \textrm{\footnotesize{swing leg length}}\\
    \textrm{\footnotesize{stance leg length}}\\
    \textrm{\footnotesize{swing leg pitch}}\\
    \textrm{\footnotesize{pelvis pitch}}\\
    \textrm{\footnotesize{swing foot pitch}}\end{pmatrix} \label{eq:walking_outputs}
\end{align}
where $\phi^y(\theta_{\textrm{tp}})$ is the ankle Cartesian pitch, and
\begin{equation}
    \psi^{st/sw}(q) = p_{hp}(q) - p_{tp}^{st/sw}(q),
\end{equation}
represents the Cartesian distance from the hip pitch joints to each of the feet. 

\begin{figure}[b!]
    \centering
    \includegraphics[width =  \columnwidth]{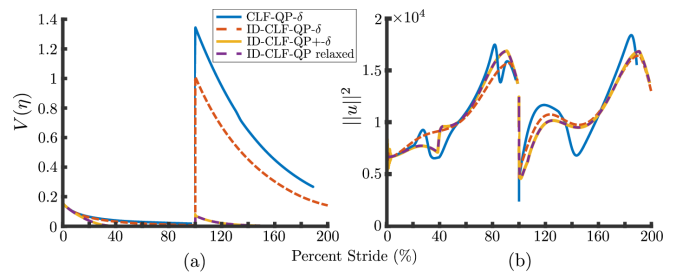} 
    \caption{Performance of the planar simulated walking gait over two steps, started from rest. Improvement is seen when the Lyapunov term is added to the cost. (a) Lyapunov function, $V(\eta)$ (b) Torque squared, $\|u\|^2$.  }
    \label{fig:simLyap}
\end{figure}

\begin{figure*}[t!]
    \vspace{4mm}
    \centering
    \includegraphics[width = \columnwidth]{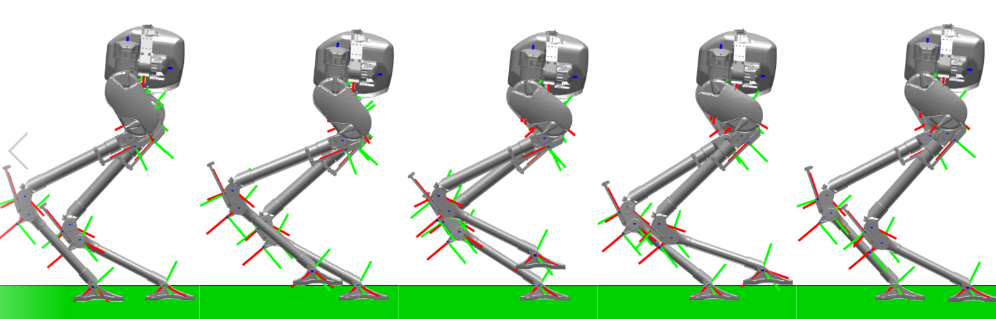}
    \includegraphics[width = \columnwidth]{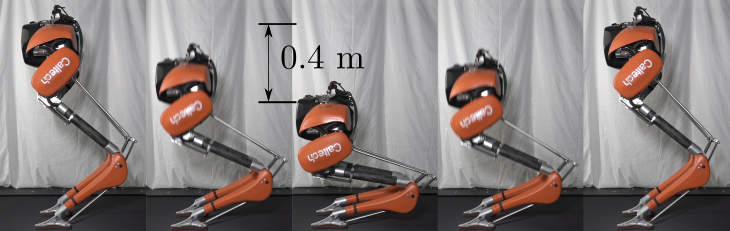}
    \caption{Time series motion tiles for simulated walking in $3$D (left) and on hardware for the crouching experiment (right).}
    \label{fig:crouchTiles}
    \vspace{-4mm}
\end{figure*}

Four controllers are then applied to the model; (\textbf{CLF-QP-$\delta$}), (\textbf{ID-CLF-QP $\delta$}), (\textbf{ID-CLF-QP+ $\delta$}), and (\textbf{ID-CLF-QP+ relaxed}). In Fig. \ref{fig:simLyap} the convergence of the Lyapunov function can be seen for a system that is perturbed to start from rest (not started on its periodic orbit) and must converge onto the periodic gait. It can be seen that the more traditionally formulated controllers do not converge quickly enough in the first step, causing an amplification of error in the second, 
 while the two cases with the Lyapunov term in the cost do. It is also interesting to note that when the Lyapunov term is in the cost, the existence of the hard convergence constraint does not significantly affect the response. While the performance differs between the four controllers, the torque applied from each is similar in magnitude and form, as can be seen in Fig. \ref{fig:simLyap}. The inverse dynamics torques are overall smoother, and the controllers with the Lyapunov derivative term in the cost have the smoothest torque profiles and best convergence performance.

\vspace{2mm} \noindent
\textbf{$3$D Compliant Walking Simulation.}
The second simulation case is a time-based walking gait on the $3$D compliant model of the robot \cite{reher2019dynamic}. For this formulation the relative degree 1 output is disregarded and four new relative degree 2 outputs - both hip yaws ($\theta_{hy}$), the swing hip roll ($\theta_{hr}$) and the floating base roll ($\phi_{r}$) - are added. The gait generated is tracking with the controllers directly on the nominal walking gait motions from an offline trajectory optimization. 

\begin{figure}[b!]
    \centering
    \includegraphics[width = \columnwidth]{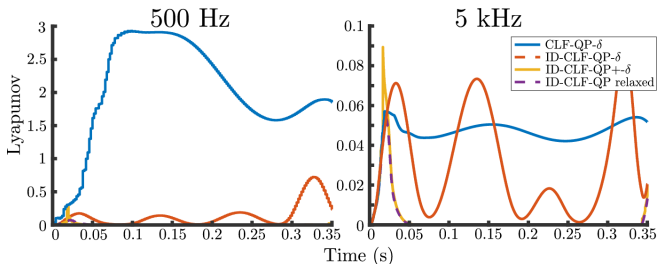}
    \vspace{-8mm}
    \caption{Lyapunov function convergence on the $3$D compliant robot for a time based step at $500$ Hz and $5$ kHz control frequencies.}
    \label{fig:sim3DLyap}
    \vspace{-2mm}
\end{figure}

\begin{figure}[b!]
    \centering
    \includegraphics[width = \columnwidth]{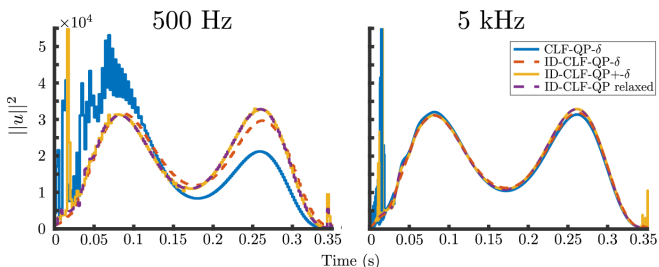}
    \vspace{-8mm}
    \caption{Torque of the $3$D compliant robot for time based step at $500$ Hz and $5$ kHz control frequencies.}
    \label{fig:sim3Dtorque}
\end{figure}

The Lyapunov function convergence and motor torques for each of the controllers can be seen in Fig. \ref{fig:sim3DLyap} and \ref{fig:sim3Dtorque}, respectively. The theory referenced in this work assumes purely continuous control, however, in reality torques are applied at a discrete intervals. We thus included the simulation results when the controllers are applied at 500 Hz and 5 kHz. An animation of the resulting simulation is also shown in \cite{papervideo}.
While the traditional (\textbf{CLF-QP-$\delta$}) and  (\textbf{ID-CLF-QP-$\delta$}) controllers see a marked degradation as loop rates decreases, the controllers which have Lyapunov derivative terms in the cost, (\textbf{ID-CLF-QP+-$\delta$}), and (\textbf{ID-CLF-QP+-relaxed}), seem minimally affected. 
Because this controller is run on the compliant model, the ODE is much more numerically stiff than in the rigid planar case. As such, we see that (\textbf{CLF-QP-$\delta$}), which uses the inverted form of the mass inertia matrix, is much more sensitive when applied at coarse frequencies.

\vspace{2mm} \noindent
\textbf{Crouching in Real-Time on Hardware.}
\begin{figure}[b!]
    \centering
    \includegraphics[width = 0.98 \columnwidth]{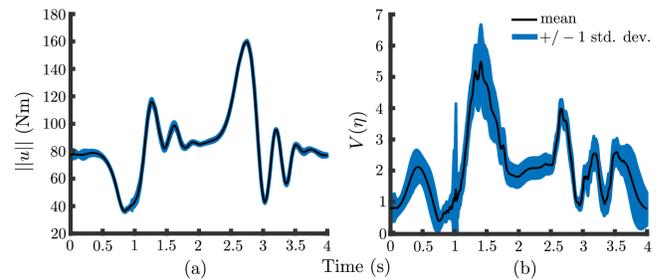}
    \vspace{-4mm}
    \caption{Torque and Lyapunov function values over $45$ crouches on hardware with the shaded areas as $+/-$ one std. deviation.}
    \label{fig:exp_torque}
    \vspace{-2mm}
\end{figure}

\begin{figure}[b!]
    \centering
    \includegraphics[width = \columnwidth]{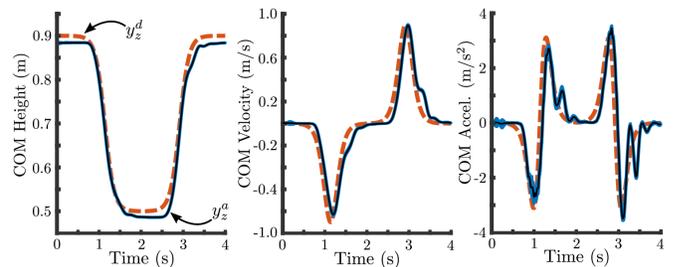}
    \vspace{-8mm}
    \caption{Height, velocity, and acceleration over $45$ crouches. Desired outputs shown as dashed, mean as black, and shaded blue as $+/-$ one std. deviation.}
    \label{fig:realTracking}
\end{figure}

\begin{figure}[t]
\vspace{4mm}
    \centering
    \includegraphics[width = \columnwidth]{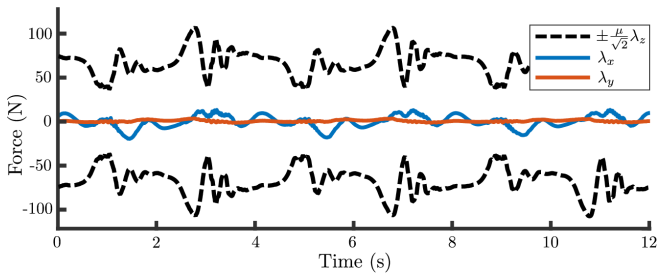}
    \caption{Contact forces for the left foot adhering to the pyramidal friction constraints over three consecutive crouches on hardware.}
    \label{fig:realFriction}
    \vspace{-4mm}
\end{figure}

Finally, the inverse dynamics motivated control Lyapunov based controller, (\textbf{ID-CLF-QP+ relaxed}),  was implemented on hardware, as Cassie went through a dynamic crouching motion (Fig. \ref{fig:crouchTiles}). The prescribed motion was a repeated crouch which moved the pelvis vertically from $0.9$ m to $0.5$ m and back, with each segment being $2$ s in duration. The source code to run a Gazebo simulation or directly implement the code on hardware is provided online \cite{papergithub}. In addition, a video of the experiment, along with animations of the previous simulations are provided in \cite{papervideo}. Six relative degree two outputs for standing were prescribed, the base positions and rotations, $y(q) = [p_b, \ \phi_b]^T$. They were then specified as high level targets on hardware. Because we are using a task-space approach, it is not necessary to encode these objectives as combinations of the actuated joint angles, and no joint level stabilization (i.e. individual joint tracking or control) was used. The controller was run on the secondary Intel NUC computer aboard Cassie, and was implemented in C++. The controller ran at a frequency of $1$ kHz, with approximately $6 \%$ timing jitter. There were only three sets of hard constraints; the dynamics as in \eqref{eq:eom}, torque bounds for each joint, and the friction constraints as in \eqref{eq:pyramid_friction} and \eqref{eq:roll}. The cost included soft constraints for the remaining holonomic constraints as well as on torque smoothness $(u_k - u_{k-1})$, in addition to the costs explicitly prescribed in (\textbf{ID-CLF-QP$^+$-relaxed}). The resulting QP was solved with the qpOASES package and had $49$ variables, and $41$ constraints. As can be seen in Fig. \ref{fig:realTracking}, the height was smoothly tracked to within several centimeters for the entirety of the motion. The norm of torque applied to all motors can be seen in Fig. \ref{fig:exp_torque}, which are smooth and satisfying all torque limitations. In addition, in Fig. \ref{fig:realFriction} the contact forces are shown to adhere to the friction cone \eqref{eq:cone_friction}.

\section{Conclusions}
\label{sec:conlcusion}
This paper presented an optimization based controller which leverages the desirable convergence results provided by control Lyapunov functions combined with implementation concepts from inverse dynamics based controllers. 
The approach was shown to be successful both in simulation and on hardware in real time. 
Further, the inclusion of a Lyapunov term in the cost helped incentive the system to converge more rapidly (as was proved in Theorem \ref{thm:main}) and improved performance with respect to discretization and model inaccuracy/stiffness.  
This was demonstrated in simulation with walking, and experimentally with crouching.   

Future work will explore improving the efficiency of the control method, with the intention of demonstrating walking in real-time on hardware. Additionally, it opens to door to providing more implementable methods for safety critical systems through control barrier functions. 
Finally, the controllers provided in this work are documented and available as part of a C++ open-source code repository, which can be used in simulation or directly on hardware \cite{papergithub}.

%%%%%%%%%% Bibliography %%%%%%%%%%%%%%%%%%%%%%%
\bibliographystyle{plain}
\bibliography{bibdata}
\end{document}